\documentclass[10pt,twocolumn,letterpaper]{article}

\usepackage{cvpr}      

\usepackage{graphicx}
\usepackage{amsmath}
\usepackage{amssymb}
\usepackage{booktabs}
\usepackage{comment}
\usepackage{xcolor}
\usepackage{soul}
\usepackage{footnote}
\usepackage[symbol]{footmisc}

\usepackage[pagebackref,breaklinks,colorlinks]{hyperref}
\usepackage[capitalize]{cleveref}
\crefname{section}{Sec.}{Secs.}
\Crefname{section}{Section}{Sections}
\Crefname{table}{Table}{Tables}
\crefname{table}{Tab.}{Tabs.}

\def\cvprPaperID{3564} 
\def\confName{CVPR}
\def\confYear{2022}
\begin{document}

\title{JRDB-Act: A Large-scale Dataset for Spatio-temporal Action, Social Group and Activity Detection}

\author{Mahsa Ehsanpour$^1$, Fatemeh Saleh$^{2, *}$, Silvio Savarese$^3$, Ian Reid$^1$, Hamid Rezatofighi$^4$ \\ 
$^1$~The University of Adelaide, $^2$~Samsung AI Cambridge, $^3$~Stanford University, $^4$~Monash University}

\maketitle
\footnotetext{$^*$~Work done while at the Australian National University~(ANU).}
\begin{abstract}
The availability of large-scale video action understanding datasets has facilitated advances in the interpretation of visual scenes containing people. However, learning to recognise human actions and their social interactions in an unconstrained real-world environment comprising numerous people, with potentially highly unbalanced and long-tailed distributed action labels from a stream of sensory data captured from a mobile robot platform remains a significant challenge, not least owing to the lack of a reflective large-scale dataset. In this paper, we introduce JRDB-Act, as an extension of the existing JRDB, which is captured by a social mobile manipulator and reflects a real distribution of human daily-life actions in a university campus environment. JRDB-Act has been densely annotated with atomic actions, comprises over 2.8M action labels, constituting a large-scale spatio-temporal action detection dataset. Each human bounding box is labeled with one pose-based action label and multiple~(optional) interaction-based action labels. Moreover JRDB-Act provides social group annotation, conducive to the task of grouping individuals based on their interactions in the scene to infer their social activities~(common activities in each social group). Each annotated label in JRDB-Act is tagged with the annotators' confidence level which contributes to the development of reliable evaluation strategies. In order to demonstrate how one can effectively utilise such annotations, we develop an end-to-end trainable pipeline to learn and infer these tasks, \textit{i.e.} individual action and social group detection. The data and the evaluation code is publicly available at \url{https://jrdb.erc.monash.edu/}.
\end{abstract}

\vspace{-.7em}
\section{Introduction}
\begin{figure}[!tbp]
  \centering
  	\includegraphics[clip,width=1.01\linewidth]{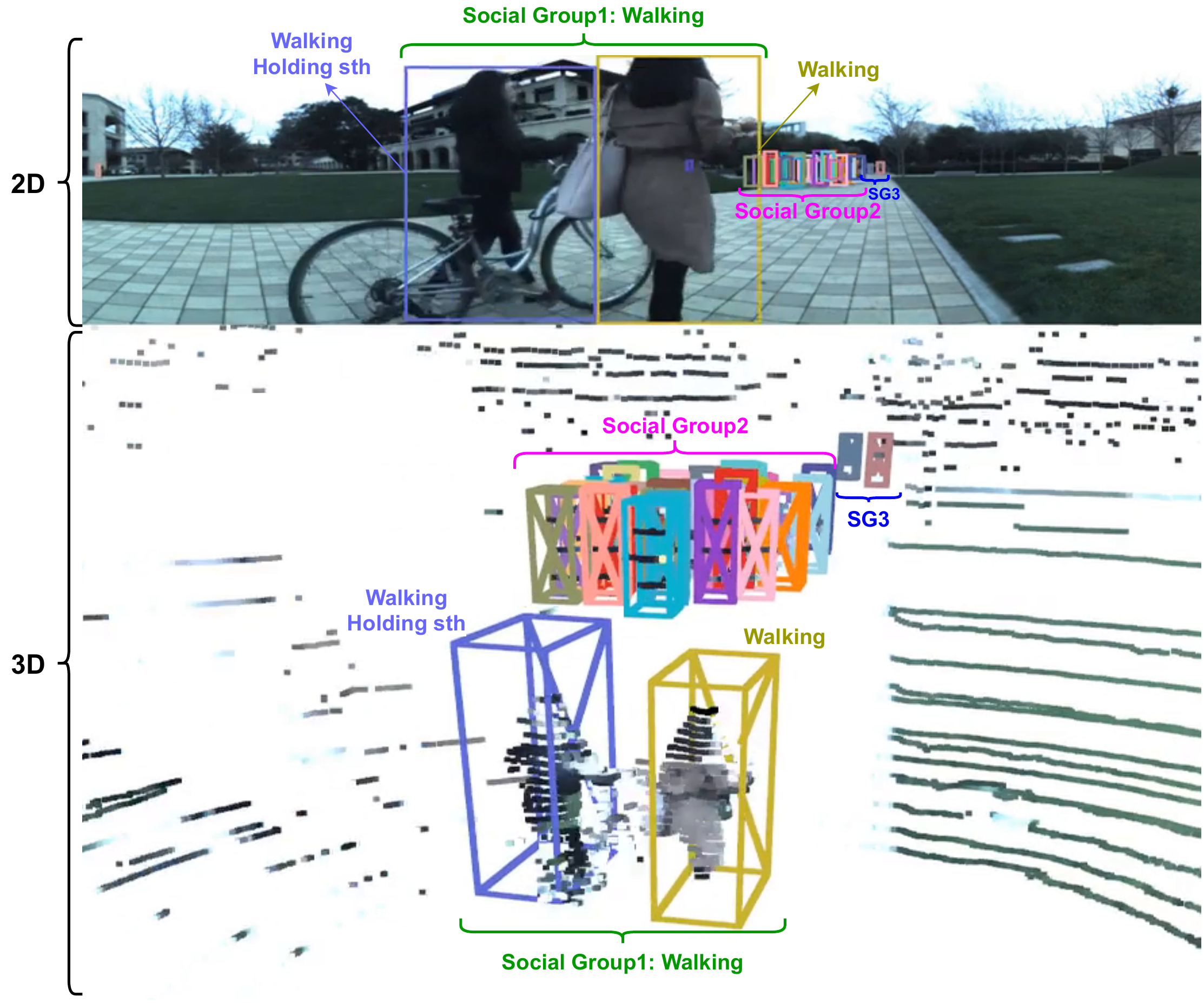}
\caption{An illustration of a single frame of the JRDB-Act dataset. As shown, the data captured with a 2D and 3D multi-modal sensory platform is accompanied with a new set of annotations including individual actions and social group formation leading to infer social activities~(common activities in each social group) to further complement the 2D and 3D detection and tracking annotation in the JRDB.}
\label{fig:teaser}
\vspace{-1.5em}
\end{figure}
\noindent
Understanding and predicting human actions and intentions are essential tasks in tackling many real-world problems such as autonomous driving, robot navigation safety, human-robot interaction, and detection of perilous behaviors in surveillance systems. Developing an AI model performing these tasks is challenging due to the high variations of human actions in an unconstrained real-world environment. Moreover, dealing with daily actions which resembles a highly unbalanced, long-tailed distribution poses new challenges for many existing approaches. 
\\
\noindent
Recently, great progress has been made to create large-scale video datasets for human activity understanding~\cite{caba2015activitynet,damen2018scaling,gu2018ava,kay2017kinetics,sigurdsson2016hollywood}.
While these popular datasets have contributed significantly to the recent advances in human activity understanding from visual data, their primary application is not targeting robotics domain and therefore rarely reflect the challenges in problems such as human-robot interaction and robot navigation in human crowded environments,~\eg shopping malls, university campus,~\etc. Such environments include not only many individuals, but also often groups of people connected to each other through some form of interaction,~\eg engaging in common activities or goals, which form the concept of social groups and activities. Moreover, in many robotics problem,~\eg for safe navigation and collision risk prediction in human environments, it is essential to anticipate every individual's action and intention way ahead of time, considering their social interactions.  To this end, the availability of a spatio-temporally dense annotated human action data is indispensable for the development and evaluation of a robotic perception system.
\\
\noindent
With this motivation, we introduce JRDB-Act, a large-scale dataset captured from a mobile robot platform, containing dense spatio-temporal individual action and social group annotation. JRDB-Act is an extension of the recently introduced JRDB~\cite{martin2019jrdb,shenoi2020jrmot}. We now elaborate the unique characteristics of JRDB-Act and our proposed method.
\\
\noindent
\textbf{New Annotations.}
We provide a set of atomic action labels for each person at each frame from the three categories of human pose, human-human, and human-object interactions, as shown in Fig.~\ref{fig:teaser}. Our action vocabulary contains common daily human actions including 11 human pose, 3 human-human, and 12 human-object interaction classes. Since these action labels are densely annotated over space and time, JRDB-Act contains over 2.8M action labels, making it one of the large-scale spatio-temporal action datasets publicly available. Furthermore, the dataset provides new unique annotations,~\ie social group labels, by assigning a group ID to each person in each frame such that individuals with the same ID represent a social group. We further provide social activity annotation for each group by inferring it from the annotated individual actions and social groups. Another novel aspect of JRDB-Act is the difficulty level annotation,~\eg easy, moderate, and difficult, for each annotated label which reflects the confidence level of annotators for the corresponding label. The provided difficulty level can be conducive to more reliable evaluation paradigms.\\
\noindent
\textbf{Unique Challenges.}
The sequences in JRDB-Act are captured from human daily-life in different indoor and outdoor places of a university campus as an unconstrained environment~\cite{martin2019jrdb} by a mobile robotic platform. Thus, they reflect the highly unbalanced distribution of human actions in real-world scenarios. Moreover, the sequences naturally include diverse levels of human population density. The average number of people per frame in JRDB-Act is 30, which is significantly higher than most popular action datasets. Further, the robot motion and the perspective view of the captured sequences makes this dataset challenging. Considering the aforementioned compelling attributes, dense annotations, and natural complexities, JRDB-Act introduces means to study new problems and challenges in human understanding for computer vision and robotics community.
\\
\noindent
\textbf{Our Proposed Method.} In order to showcase the potential research directions and challenges required to be tackled in JRDB-Act, we develop an end-to-end trainable pipeline for both individual action and social group detection tasks. Our method uses the panoramic video clips as input and adopts a similar backbone as~\cite{ehsanpour2020joint} to extract spatio-temporal individual features. However, we fuse additional pair-wise geometrical features and incorporate a novel eigenvalue-based loss function to improve the social group detection performance compared to~\cite{ehsanpour2020joint}. We also suggest a simple, yet effective strategy to handle the unbalanced nature of action labels by partitioning and balancing action loss functions based on the occurring frequency of action classes in the dataset. 
\section{Related Work}
\noindent
\textbf{Datasets.}
Over the last decade, multiple video action datasets such as HMDB-51~\cite{kuehne2011hmdb} and UCF101~\cite{soomro2012ucf101} have been introduced which consist of short clips for the video classification task~\cite{gorelick2007actions,marszalek2009actions,schuldt2004recognizing}. Since these datasets are not large and diverse enough to train deep models, large-scale video datasets such as Sports1M~\cite{karpathy2014large}, YouTube-8M~\cite{abu2016youtube}, Something-something~\cite{goyal2017something} and Kinetics~\cite{kay2017kinetics} have been introduced for the task of video action classification. Some other video datasets such as ActivityNet~\cite{caba2015activitynet}, THUMOS~\cite{idrees2017thumos}, 
MultiTHUMOS~\cite{yeung2018every}, Charades~\cite{sigurdsson2016hollywood} and HACS~\cite{zhao2019hacs} contain untrimmed videos for the task of temporal action localization. Few datasets, such as CMU~\cite{ke2005efficient}, MSR Actions~\cite{yuan2009discriminative}, UCF Sports~\cite{rodriguez2008action} and JHMDB~\cite{jhuang2013towards} provide spatial as well as temporal localization. The small number of action categories and the limited number of short video clips motivated the community to introduce AVA~\cite{gu2018ava} and AVA-Kinetics~\cite{li2020avakinetics}, two large-scale spatio-temporal action detection datasets. In AVA, spatio-temporal action labels are provided for one frame per second, in which every person is annotated with a bounding box and at least one action. The AVA-Kinetics dataset extends Kinetics with AVA-style annotation.
There are also a number of video datasets such as SOA~\cite{ray2018scenes} and HVU~\cite{diba2020large} that provide multi-label annotation by providing scene, object, event, attribute and concept labels in a video, still limited to the video classification task. As another group of datasets, instructional video analysis datasets~\cite{ben2020ikea, damen2018scaling, rohrbach2012database,tang2019coin} have been released which are focused on a specific domain such as cooking or furniture assembly. Volleyball~\cite{ibrahim2016hierarchical} and Collective Activity Dataset~(CAD)~\cite{choi2009they} have been introduced with a focus on group activity recognition. In these datasets, actors are annotated with an action label and the whole scene is annotated with one group activity label. However, a real scene generally comprises several groups of people with potentially different social activities. Recently, CAD has been extended in terms of annotations to Social-CAD~\cite{ehsanpour2020joint}, in which different social groups and their corresponding social activities have been annotated. While Social-CAD is the first attempt to tackle spatio-temporal action and social group detection tasks, it only contains 44 sequences with limited labels. Although all these datasets have vastly contributed to the recent advances in human action understanding in videos, they are not capable of reflecting challenges in robotics applications in human crowded environments. To target such specific application domains~\eg social robot navigation and human-robot interaction, we propose JRDB-Act, a large-scale spatio-temporal human action, social group and per-group social activity detection dataset captured from a mobile robot which has been annotated densely in space and time.
\\
\noindent
\textbf{Action Analysis Frameworks.} Over the past few years, there have been extensive studies on video classification~\cite{carreira2017quo, donahue2015long, simonyan2014two, yue2015beyond} and temporal action detection~\cite{sigurdsson2017asynchronous,sigurdsson2016hollywood,xu2017r,zhou2018temporal}. Recently, by introducing spatio-temporally annotated datasets such as AVA~\cite{gu2018ava}, the spatio-temporal action detection task~\cite{feichtenhofer2019slowfast,girdhar2018better,girdhar2019video,li2018recurrent,sun2018actor,wu2019long} has received considerable attention.
In parallel, there also exist works focusing on group activity recognition on datasets,~\eg Volleyball~\cite{ibrahim2016hierarchical} and CAD~\cite{choi2009they}, where the aim is to predict a single group activity label for the entire scene~\cite{choi2012unified, choi2013understanding, choi2011learning, li2021groupformer, li2018did}.
Although these approaches try to recognise the interactions between people for group activity recognition, they are not capable of inferring social groups. Recently, CAD~\cite{choi2009they} has been augmented with social group and social activity label per group in~\cite{ehsanpour2020joint} and a corresponding framework is proposed to detect individuals' action, social groups, and social activity for each group in the scene. 
However, as substantiated by our experiments, this framework does not generalize well to the natural complexities of JRDB-Act. To improve upon this framework's performance, we \emph{(i)} exploit the bounding box locations to derive pair-wise geometrical features and further incorporate an eigenvalue-based loss to enhance the social group detection task and, \emph{(ii)} suggest a simple strategy,~\ie a loss partitioning approach, to handle the unbalanced nature of action labels in the dataset.
\vspace{-0.7em}

\section{The JRDB-Act Dataset}
\noindent
The multi-modal JRDB dataset~\cite{martin2019jrdb} is composed of 64 minutes of sensory data captured by the mobile JackRabbot robot, containing 54 sequences of indoor and outdoor scenes in a university campus environment, covering different human poses and social interactions. JRDB provides 1)~over 2.4 million 2D bounding boxes for all the people visible in the five stereo RGB cameras, capturing a panoramic cylindrical 360\textdegree  image view, 2)~over 1.8 million 3D oriented bounding boxes in the point-clouds captured from the two 16-array LiDAR sensors, 3)~association of all the 3D bounding boxes with the corresponding 2D boxes, and 4)~track ID of all the 2D and 3D boxes over time. While the provided annotations are useful for human localization and tracking, JRDB lacks sufficient information for social human activity analysis. Therefore we propose JRDB-Act by providing additional individuals' human action and social group annotation on top of the existing JRDB. All these annotations make JRDB-Act the only available dataset for multi-task learning of human detection, tracking, individual action, social group and per-group social activity detection. JRDB-Act is manually annotated by a group of annotators, instructed for each task to ensure consistency over the dataset. Then, it has been inspected by another group, instructed for the quality assessment of the provided annotations.
The rest of this section provides details of JRDB-Act regarding annotations, benchmarking, and statistics.

\begin{figure*}[t]\vspace{-1em}
  \centering
		\includegraphics[clip,width=1\linewidth]{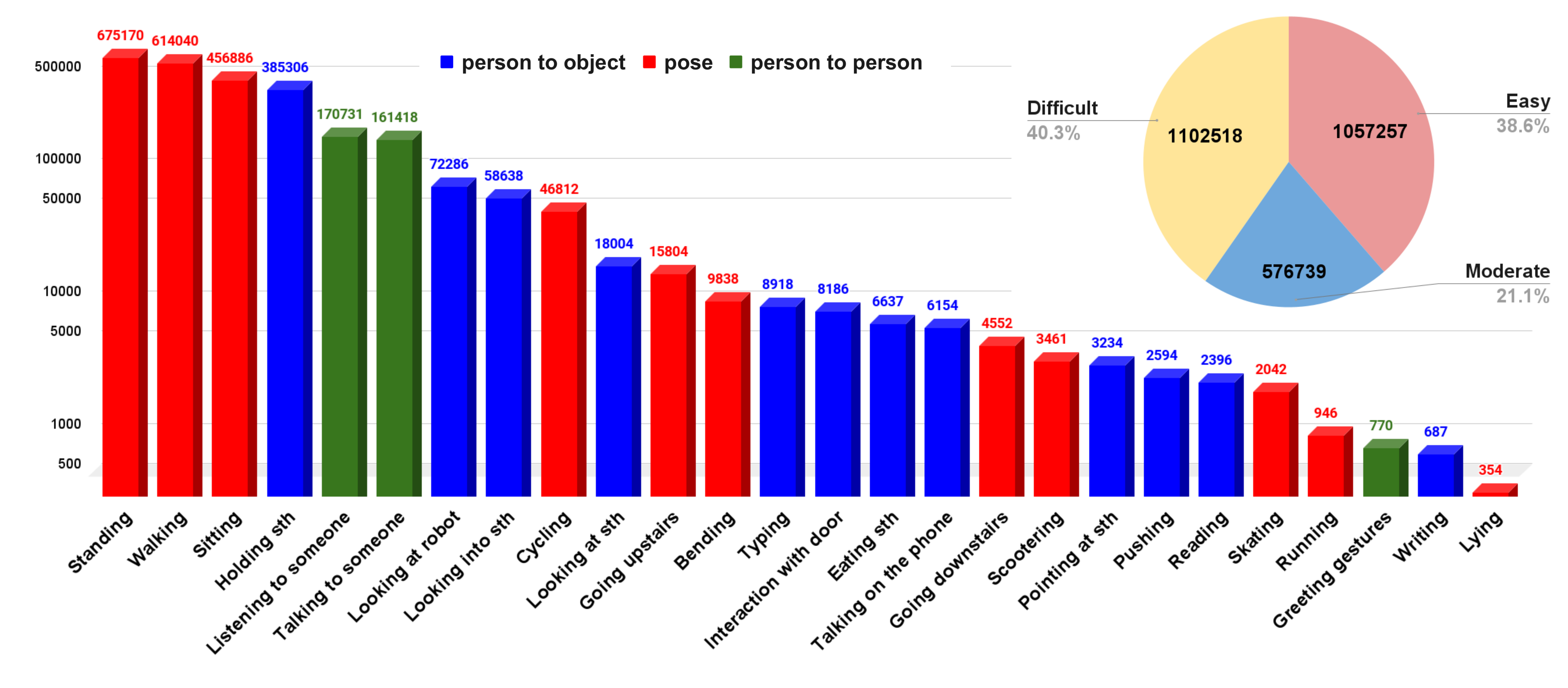}
		\vspace{-1.5em}
\caption{Left: The distribution of action classes in {\bf \emph{log-scale}} sorted by descending order, with colors indicating action types. Right: The distribution of different difficulty levels in action label annotations.}
\label{action_distribution}
\end{figure*}

\noindent
\textbf{A.~Action Vocabulary. }Since JRDB is collected in a university campus environment, our action vocabulary includes common daily human actions. Through a comprehensive inspection of the dataset, we concluded 11 pose-based, 3 human-human interaction and 12 human-object interaction action labels. Fig.~\ref{action_distribution} demonstrates the list of existing action labels per category in JRDB-Act.  
\begin{figure*}[t]
\vspace{-5pt}
  \centering
		\includegraphics[clip,width=1\linewidth]{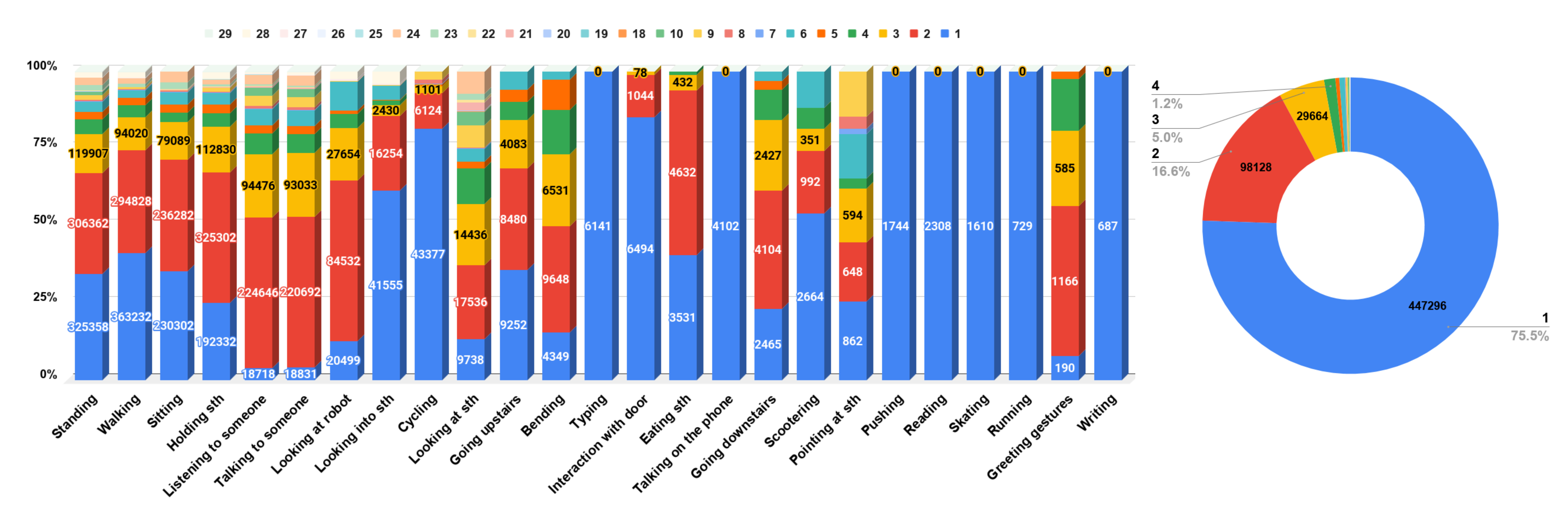}\vspace{-0.5em}
\caption{Left: The distribution of social group size for each social activity label shown in individual bars, with colors indicating the size. Right: The distribution of social group size for the entire dataset.}
\label{social-activity}\vspace{-1.5em}
\end{figure*}

\noindent
\textbf{B.~Action Annotation.} Action annotation is densely provided per-frame~(7 fps) and per-box for both the LiDAR and video sequences. However, the panoramic videos are used to annotate the action labels. During the annotation process, we utilised JRDB annotated 2D-bounding boxes and track-IDs; for each bounding box, one~(mandatory) pose-based and an arbitrary number of~(optional) interaction-based action labels were selected from the available list of action vocabulary. If none of the classes in the list were descriptive for a bounding box, annotators were able to tag the box as \textit{miscellaneous-[description]} for each label category, and the descriptions were later used to expand the action vocabulary with the newly discovered labels. Annotators also tagged each action label with its corresponding difficulty level indicating the annotator's confidence level for the corresponding label. There are scenarios where 1)~action label is obvious, 2)~there is uncertainty in the action label but we can take a probable guess, and 3)~the person is far away from the camera or occluded, however the action could be inferred from some evidences such as its past history and its current movement. We respectively tag these scenarios as \textit{easy, moderate,} and \textit{difficult}. In some cases, it is not possible to infer the action as the bounding box is fully-occluded in the duration of the video or the person is very far from the robot. Here, both the pose-based action label and its corresponding difficulty level are tagged as \textit{impossible}. The provided difficulty level can be conducive to more reliable and fair evaluation protocols.

\noindent
\textbf{C.~Social Group Annotation.} People in a scene may form different social groups~\cite{ehsanpour2020joint}, while each group is engaged with a social activity. To provide group annotation, a unique group ID is assigned to those belonging to the same social group in each frame and this assignment may vary over time. Each group label is tagged with a difficulty level to reflect the annotator's confidence level. We used \textit{easy, moderate}, and \textit{difficult} for the cases where the group membership was respectively 1)~easily recognisable, 2)~could be estimated based on some visual and temporal cues, 3)~could not be inferred due to the distance from the camera or occlusion. Given the annotated social groups and individuals' action labels in each frame, we generated a \emph{pseudo groundtruth} social activity label for each group using the most frequent individual action labels in that group. We also assign a difficulty level to the inferred social activity labels by averaging the corresponding individual actions' difficulty levels.

\noindent
\textbf{D.~JRDB-Act splits. }Following JRDB splits, JRDB-Act is divided into training, validation, and test sets at video level, thus, all the frames of a video sequence appear in one specific split. The 54 video sequences are split into 20 training, 7 validation, and 27 test videos. For the purpose of consistency with the standard evaluation of other relevant datasets, we evaluate all the task on the key-frames, which are sampled every one second, resulting in 1419 training, 404 validation, and 1802 test samples. 

\noindent
\textbf{E.~Benchmark and Metrics.} Our evaluation is performed on the key-frame level following the standard practice in~\cite{gu2018ava}. We adopt the widely used average precision~(AP) using an IoU threshold of 0.5, following the standard PASCAL~VOC~\cite{everingham2015pascal} challenge, and customize it to report the performance of each task. To report the performance of social grouping on a set of detected bounding boxes, we first calculate a list of true positive boxes~(TP) for each detection confidence threshold. Then, similar to~\cite{ehsanpour2020joint,xie2016unsupervised}, we determine a correspondence between the predicted and truth group IDs by solving an ID assignment between the refined prediction~(TP) list and the groundtruth list. Finally we re-calculate the final number of true positives considering the group IDs and use AP to report the final results. Mean AP~(mAP) is also used to report the performance of individual action and social activity detection tasks, following the same practice as in~\cite{gu2018ava}. See supp. material for a comprehensive explanation of our evaluation strategy.

\noindent
\textbf{F.~JRDB-Act Statistics. }Fig.~\ref{action_distribution} shows the JRDB-Act's distribution of annotated individuals' action labels in log-scale representing a long-tailed distribution in the dataset. Further, in the pie chart of Fig.~\ref{action_distribution}, difficulty level distribution in action labels is reflected in which only 61.4\% of action labels are annotated with respect to the visual cues~(tagged as \textit{easy} and \textit{moderate}) and the remaining 38.6\% are inferred based on the bounding box history or movement~(tagged as \textit{difficult}).
Fig.~\ref{social-activity} demonstrates the distribution of social activity labels with respect to the size of social groups. The donut chart in Fig.~\ref{social-activity} indicates the distribution of social group size in the dataset. As illustrated, 75.5\%, 16.6\%, 5\%, 1.2\%  of social groups consist of one, two, three, and four members respectively and only 1\% of the data contain groups with five or more members~(maximum 29 members).
\vspace{-0.7em}

\section{Proposed Baseline}\label{method}
\begin{figure*}[h]\vspace{-.7em}
 \centering
	\includegraphics[clip,width=1\linewidth]{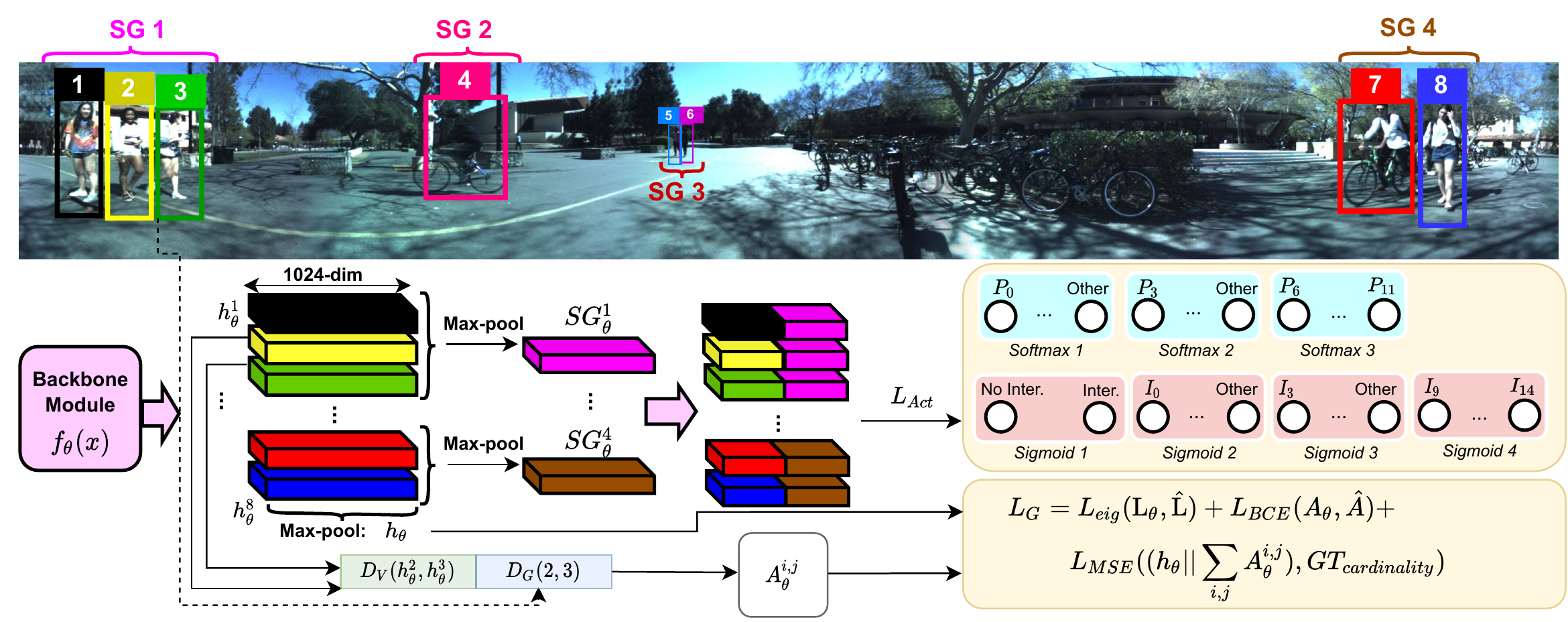}
\vspace{-1em}
\caption{Overview of our framework during training. Given the spatio-temporal feature representation of the individuals denoted by $h_{\theta}^i$ in the key-frame, we optimize two tasks. First, to learn the individual actions, we compute the individual's feature map by concatenating the individual's visual feature and its corresponding social group's feature map ($SG_{\theta}^i$) obtained by max-pooling the feature maps of its members. Then, to compute $L_{Act}$, we compute cross entropy and binary cross entropy losses for each pose-based~($P$) and interaction-based~($I$) action groups. Second, to learn the social group formation and the social group cardinality, we calculate the similarity matrix $A_{\theta}$ between individuals based on their pair-wise geometrical~($D_G(i,j)$) and feature distance extracted from the backbone~($D_V(h_{\theta}^i,h_{\theta}^j)$) and utilise it along with the extracted spatio-temporal feature~($h_{\theta}$) to compute different loss terms as in $L_G$.}  
\label{baseline}
\vspace{-5pt}
\end{figure*}
\begin{figure}[bt]\vspace{-.7em}
  \centering
		\includegraphics[clip,width=1\linewidth]{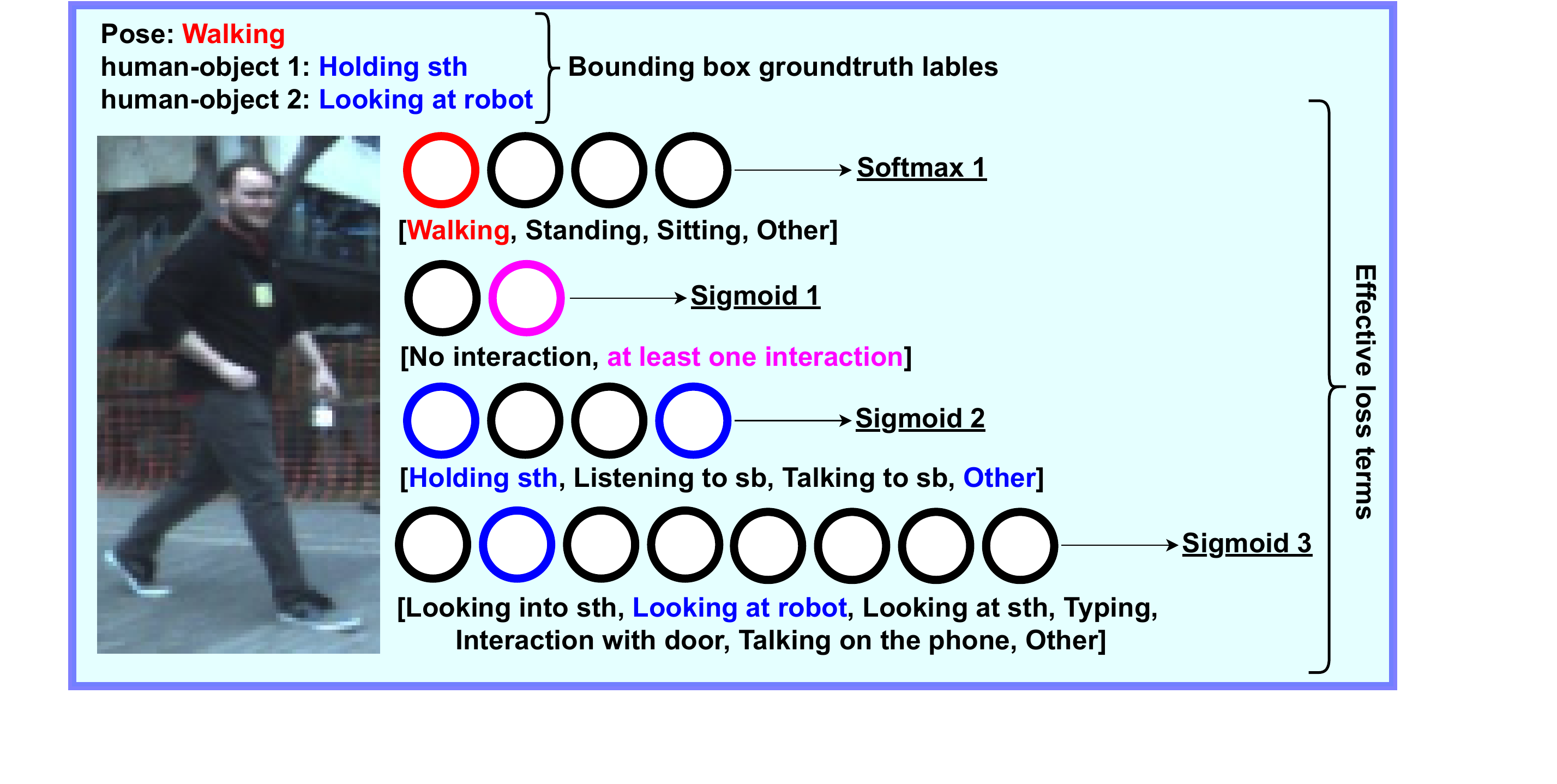}
		\vspace{-1.9em}
\caption{Illustration of different softmax and sigmoid terms of $L_{Act}$ for a training sample. As shown, there are 3 groundtruth actions in this sample including one from the pose-based and two from the human-object interaction categories. For the pose-based action, only one softmax is activated as ``Walking'' belongs to ``Softmax 1''. The first sigmoid determines whether there is an interaction-based action. The subsequent sigmoids specifically determine the present interaction-based action labels. Here, ``Holding sth'' belongs to ``Sigmoid 2'' and ``Looking at robot'' falls into the ``Other''. Thus, the third sigmoid is activated to recognise the ``Looking at robot'' action.}
\vspace{-20pt}
\label{fig:L_ACT}
\end{figure}

\noindent
We propose an end-to-end trainable baseline for spatio-temporal detection of individuals' actions, social groups, and social activities per group in videos. The architecture of our model is illustrated in Fig.~\ref{baseline}. We utilise the same backbone $f_{\theta}(x)$ as in~\cite{ehsanpour2020joint} including the I3D feature extractor, the self-attention, and the graph attention modules to extract rich spatio-temporal feature map for each individual in which social interactions are encoded. To further enhance the social grouping performance and to reduce the discrepancy between train and inference compared to~\cite{ehsanpour2020joint}, we propose to incorporate an eigenvalue-based loss function~\cite{dang2018eigendecomposition} on the similarity matrix extracted from the visual features and geometrical relations between the detected bounding boxes. Further, in order to overcome the highly unbalanced nature of action labels in the data, we propose to utilise softmax/sigmoid loss partitioning approach inspired by~\cite{li2020overcoming}. 
\\
\noindent
\textbf{Learning Social Group Formation.}\label{sec:LSGF}
Social groups in a scene can be shown as a graph in which nodes are the individuals and the edges indicate the connectivities between them. The graph of the groundtruth social groups can be presented by a matrix $\hat A$ consisting of 0 and 1 in which $\hat A_{i,j}$ indicates whether the pair $(i,j)$ belongs to the same social group. $A_{\theta}$ is formed by the model in which for each pair of bounding boxes $i$ and $j$, the normalised GIoU~\cite{rezatofighi2019generalized}, $D_G(i,j)$, representing a geometrical similarity between each pair is calculated such that 0 and 1 represent far and close boxes, respectively. The normalised similarity between the visual features~(extracted from $f_{\theta}(x)$) of two bounding boxes $i$ and $j$ is also calculated as $D_V(h^{i}_{\theta},h^{j}_{\theta})$. The final $A_{\theta}^{i,j}$ is then attained by the concatenation of $D_V(h^{i}_{\theta},h^{j}_{\theta})$ and $D_G(i,j)$ and utilising a MLP layer to project the 2-dim vector to a 1-dim vector. The training objective in learning social groups is to reduce the discrepancy between the predicted $A_{\theta}$ and $\hat A$.
To this end, we utilise a binary cross entropy loss between the elements of $A_{\theta}$ and $\hat A$ denoted by $L_{BCE}$ in Eq.~\ref{eq:L_G}. Further, since the number of connected components~(social groups) in the groundtruth matrix $\hat A$ is equal to the number of zero eigenvalues of its laplacian matrix ${\hat L}$, we want the laplacian matrix of $A_{\theta}$ denoted by ${L}_{\theta}$ to have the same number of zero eigenvalues as in ${\hat L}$. To this end, we utilise $L_{eig}(\theta)$ denoted by Eq.~\ref{eq:LG},
\begin{equation}\label{eq:LG}
L_{eig}(\theta)=\hat e^{T}{L}_{\theta}^{T}{L}_{\theta}\hat e+\alpha\exp(-\beta tr(\bar{{L}}_{\theta}^{T}\bar{{L}}_{\theta}))\vspace{-.5em} 
\end{equation}
in which $\hat e$ is the groundtruth eigenvector corresponding to the zero eigenvalue, ${L}_{\theta}$ is the laplacian matrix corresponding to the predicted similiraty matrix $A_{\theta}$ and $\alpha$ and $\beta$ are coefficients. The proof of Eq.~\ref{eq:LG} is stated in the supp. material. 
The loss in Eq.~\ref{eq:LG} is inspired by the fully differentiable, eigendecomposition-free loss proposed in~\cite{dang2018eigendecomposition} to train a deep network whose loss depends on the eigenvector corresponding to the single zero eigenvalue of a matrix predicted by the network. We extend it to our problem with multiple zero eigenvalues indicating the number of social groups. To learn the number of social groups, as a cardinality loss, we utilise a mean square error function between the groundtruth number of social groups and the 1-dim learned feature from the concatenation of $h_{\theta}$~(max-pool of boxes' visual features) and the summation of the $A_{\theta}$ elements denoted by $L_{MSE}$ in Eq.~\ref{eq:L_G}.
\vspace{-.6em}
\begin{equation}\label{eq:L_G}
\begin{aligned}
    L_G=L_{BCE}(A_{\theta},\hat A)+L_{eig}({L}_{\theta},{\hat L})+\\L_{MSE}((h_{\theta}||\sum_{i}A_{\theta}^{i}),GT_{cardinality})
\end{aligned}
\vspace{-1em}
\end{equation}
\\
\noindent
\textbf{Learning Actions.}\label{sec:LA}
Each bounding box is annotated with one pose-based and an arbitrary number of interaction-based action labels and the occurrence of action classes is highly unbalanced in the dataset. One naive way to learn actions is to use a cross entropy loss to learn pose-based and a binary cross entropy loss to learn interaction-based actions. However, we empirically observe that action classifier's performance is highly harmed by the unbalanced nature of action labels. To overcome this problem, we divide the pose-based and interaction-based action classes into several disjoint partitions. The number of samples of the least frequent class in each partition, is greater than 0.1 of the number of samples of the highest frequent class in that partition. In each partition excluding the last one, we add an ``Other'' class which shows the presence of an action class in the less frequent partitions. We have 3 and 4 partitions for pose-based and interaction-based partitions respectively. The list of action labels in each partition is provided in the supp. material. We then train each pose-based and interaction-based partition separately by using cross entropy and binary cross entropy losses respectively as in Eq.~\ref{eq:action}. Further, to maintain the balance, we only train partitions with an existing groundtruth label for each training sample. An illustration of our action learning strategy is shown in Fig.~\ref{fig:L_ACT}.
\begin{equation}\label{eq:action}
\begin{aligned}
    L_{Act}=\sum_{i=0}^{2}\lambda_i L_{CE}(P^{i}_{\theta},P^{i})+\sum_{j=0}^{3}\lambda_j L_{BCE}(I^j_{\theta},I^j)
\end{aligned}
\vspace{-.7em}
\end{equation}
In Eq.~\ref{eq:action}$, \lambda$ is a coefficient, $P_{\theta}^{i}$ and $I_{\theta}^{j}$ are the predicted pose-based and interaction-based actions, and $P^{i}$ and $I^{j}$ are the corresponding groundtruth labels respectively.
\\
\begin{table*}[h]
\scriptsize
\centering
\begin{tabular}{c| c c c| c c c c c c c c c}
\hline
Method & grouping loss & Cardinality & Geo feature & G1 AP$\uparrow$ & G2 AP$\uparrow$ & G3 AP$\uparrow$ & G4 AP$\uparrow$ & G5$^+$ AP$\uparrow$ &overall AP$\uparrow$ \\ 
\hline\hline
Baseline1~\cite{ehsanpour2020joint} &BCE &H &- &8.0 &29.3 &37.5 & \bf 65.4 &\bf 67.0 &41.4  \\
\hline
Baseline2 &BCE &H &\checkmark &26.1 &57.0 &\bf61.2 &63.0 &53.7 &52.2 \\
\hline
Baseline3 &BCE &MSE &\checkmark & 79.6 &63.0 &43.7 &56.9 &40.7 &56.8 \\
\hline\hline
\bf Ours &BCE+EIGEN &MSE &\checkmark &\bf81.4 &\bf 64.8 & 49.1 & 63.2 &37.2 &\bf 59.2 \\
\hline
\end{tabular}
\caption{Social grouping ablation study on JRDB-Act validation-set using groundtruth bounding boxes. G1, G2, G3, G4, G5$^+$ indicate social groups with 1, 2, 3, 4, 5 or more members.}
\label{tab:G_ab}
\end{table*}
\begin{table}[h]
\scriptsize
\centering
\begin{tabular}{c |c}
\hline
Method &Action mAP$\uparrow$ \\ 
\hline\hline
[CE+BCE] &8.0  \\
\hline
[W-CE+W-BCE] &8.1  \\
\hline
\bf [M-CE+M-BCE]~[Ours] & \bf 9.0  \\
\hline
\end{tabular}
\caption{Individual action detection ablation study on JRDB-Act validation-set using groundtruth bounding boxes.}
\label{Tab:A_ab}
\end{table}
\noindent
\textbf{Training.} Our model takes as input a video clip with the key-frame located at the end. The input clip is then fed to the backbone to obtain spatio-temporal feature map of the individuals in the key-frame denoted by $h_{\theta}^i$. The similarity matrix $A_{\theta}$ between individuals is calculated based on their pair-wise geometrical and feature distance. The calculated similarity matrix and the extracted spatio-temporal features are then utilised to learn the social grouping loss $L_G$ denoted by Eq.~\ref{eq:L_G}. Given the groundtruth social connections in training, we obtain each social group's feature map by max-pooling the features of its members. Each individual's feature representation is concatenated with its social group feature map. Individual's obtained feature map are utilised to learn the action loss $L_{Act}$ as in Eq.~\ref{eq:action}. As shown in Fig.~\ref{fig:L_ACT}, For each training sample we only activate the terms of $L_{Act}$ in which there exists a groundtruth label and set the other terms to zero to avoid training with groundtruth vectors of all zeros. The total training objective is stated in Eq.~\ref{eq:total}.\vspace{-.7em}
\begin{equation}\label{eq:total}
    L_{total}=L_G+L_{Act}\vspace{-1em}
\end{equation}
\\
\noindent
\textbf{Inference.}~At test time, for individual action prediction, we perform softmax operation on the predictions of each cross entropy and sigmoid operation on predictions of each binary cross entropy functions. We then choose the predicted action labels based on a hierarchical approach starting from the first partition and going to the next one in the hierarchy only if the ``Other'' class is predicted. For social group prediction, we perform graph spectral clustering~\cite{zelnik2005self} on the obtained similarity matrix between individuals and by utilising the predicted number of social groups. Since the social activity label of each group is the most frequent action labels of its members, we follow the same strategy and infer the activity of each predicted social group from the predicted action labels of its individuals.
\begin{table*}[h]
\scriptsize
\centering
\begin{tabular}{ c|c c c c c c | c| c c}
\hline
Method &G1 AP$\uparrow$ &G2 AP$\uparrow$ &G3 AP$\uparrow$ &G4 AP$\uparrow$ &G5$^+$ AP$\uparrow$   &overall AP$\uparrow$  &Action mAP$\uparrow$ & G-Act mAP1$\uparrow$   & G-Act mAP2$\uparrow$\\ 
\hline\hline
\cite{ehsanpour2020joint}+Faster-RCNN &9.5 &24.3 &21.2 &\bf 39.8 &10.8 &21.1 &4.4 &3.5 &1.3 \\
\hline
\cite{ehsanpour2020joint}+MMPAT &11.8 &27.5 &22.4 &38.8 &\bf 24.6 &25.0 &4.9 &3.5 &1.3 \\
\hline \hline
Ours+Faster-RCNN  &42.5 &\bf 40.8 &23.1 &25.6 &13.4 &29.1 &5.3 & 4.4 &3.4\\
\hline
Ours+MMPAT  &\bf 56.6 &39.5 &\bf 24.3 &22.4 &14.8 &\bf31.5 &\bf 5.4 &\bf 4.7 &\bf 3.4\\
\hline
\end{tabular}
\caption{Final results of our model against~\cite{ehsanpour2020joint} on JRDB-Act test-set using two different sets of detection bounding boxes~(Faster-RCNN~\cite{ren2016faster} and MMPAT~\cite{he2021know}) and by considering labels with Easy and Moderate difficulty tags in evaluation.}
\label{tab:final}
\vspace{-0.7em}
\end{table*}
\begin{table*}[h]
\scriptsize
\centering
\begin{tabular}{c| c c c c c c | c|c c}
\hline
[Ours] &G1$\uparrow$ &G2$\uparrow$ &G3$\uparrow$ &G4$\uparrow$ &G5$^+$$\uparrow$   &overall AP$\uparrow$  &Action mAP$\uparrow$   &G-Act mAP1$\uparrow$ &G-Act mAP2$\uparrow$\\ 
\hline\hline
[E,M,D] &34.9 &37.3 &18.3 &16.4 &7.6 &22.9 &4.4 &3.5 &2.7\\
\hline
[E,M] &42.5 &40.8 &23.1 &25.6 &13.4 &29.1 &5.3 &\bf 4.4 &3.4\\
\hline
[E] &\bf 44.4 &\bf 42.7 &\bf 27.1 &\bf 28.4 &\bf 13.9 &\bf 31.3 &\bf 5.7 &\bf 4.4 &\bf 3.5\\
\hline
\end{tabular}
\caption{E:Easy, M: Moderate and D:Difficult. Performance of different tasks wrt the difficulty tag on JRDB-Act test-set.}
\label{tab:diff}
\vspace{-1em}
\end{table*}
\section{Experiments}\label{sec:exp}
\noindent
In this section, we provide implementation details of our framework,
evaluate different aspects of it, and present a comparison against the existing method proposed in ~\cite{ehsanpour2020joint}.
\\
\noindent
{\bf Implementation Details.}
The backbone setup and hyper-parameters are identical to~\cite{ehsanpour2020joint}. We utilize ADAM optimizer with $\beta_{1}=0.9, \beta_{2}=0.999, \epsilon=10^{-8}$. $\alpha$ and $\beta$ in Eq.~\ref{eq:LG} are set to 1. Since the training objective includes learning social groups and actions and to effectively learn both tasks, we train the model in two stages. In the first stage, we train the model with $L_G$ for 50 epochs with a mini-batch size of 1 and an initial learning rate of $10^{-4}$. We then fine-tune the network with $L_{total}$ for 50 epochs. The learning rate is reduced by a factor of $10^{-1}$ on validation loss plateau. Input video clips to the model are 15 frames long with the annotated key-frame at the end. See the supp. material for more implementation details.  
\\
\noindent
{\bf Ablation Studies.} All the ablations in Tab.~\ref{tab:G_ab} and Tab.~\ref{Tab:A_ab} are performed using groundtruth bounding boxes on validation-set to remove the effect of detection performance from the experiments. Further, evaluation for each task is performed by considering the corresponding groundtruth labels with easy and moderate difficulty tags and difficult labels are removed from the evaluation. Labels with difficult tags however are used for evaluation on test-set in Tab.~\ref{tab:diff}.\\    
\noindent
{\bf A.~Social Group Formation:} We compare our framework against three baselines for predicting social groups in terms of grouping AP for groups with different number of members and the average of obtained grouping APs. Within our suggested framework, the network learns to estimate the number of social groups by minimizing a mean square error loss during training; we indicate this by $MSE$ in the cardinality column in Tab.~\ref{tab:G_ab}. On the contrary, the graph clustering approach used in~\cite{ehsanpour2020joint} requires the number of social groups to be known in advance and thus, relies on a heuristic~\cite{ng2002spectral} to infer this number; we indicate this with $H$ in the cardinality column. Accordingly, we define three baselines in Tab.~\ref{tab:G_ab}. [Baseline1]~\cite{ehsanpour2020joint}, addresses the group formation task with grouping loss consists of a single binary cross entropy based on the visual features of individuals, indicated by BCE in the grouping loss column, and the graph spectral clustering utilizes the heuristic to infer the number of social groups. As validated by our experiments, this heuristic underestimates the number of groups; \textit{i.e.,} spectral clustering tends to group everyone into few or even a single group. Thus, the performance of this baseline for groups with sizes 4, 5 and above is fortuitously better than the other methods while it performs significantly worse on the lower group size categories. [Baseline2] extends [Baseline1] by exploiting geometrical features in addition to the visual features. Evidently, the geometrical features lead to better performance in identification of small-sized social groups. Similarly, [Baseline3] extends [Baseline2] by learning the social group cardinality instead of adopting the heuristic; this significantly boosts the group formation performance for small-sized groups. Finally, [Ours] shows the effect of utilizing the eigen-value loss in our framework which yields the highest overall group formation results. 
\\
\noindent
{\bf B.~Action and Social Activity Prediction:} We show the effectiveness of our proposed strategy~(\ie loss partitioning) to deal with highly unbalanced individuals' action labels in Tab.~\ref{Tab:A_ab}. [Baseline1], utilizes a single cross entropy loss and a single binary cross entropy loss~[CE+BCE] to learn pose-based and interaction-based action classes respectively. [Baseline2]~\cite{ehsanpour2020joint}, utilizes the cross entropy loss and the binary cross entropy loss functions in a weighted manner~[W-CE+W-BCE]. Normalized weights of action labels is calculated based on the inverse of their occurrence frequency in train and validation sets. Finally, in [Ours], we utilize the loss partitioning strategy using multiple cross entropy and binary cross entropy losses~[M-CE+M-BCE] as elaborated in Sec.~\ref{method}.
As validated by our experiments, weighting strategy does not address the unbalanced distribution of action classes in the data, whereas the proposed loss partitioning approach shows improvement compared to the baselines. Finally, social activity labels are inferred from the predicted social groups and individuals' actions as the most frequent actions performed by the members of that group. Social activity labels are evaluated by ignoring social groups indicated by the G-Act mAP1 column~(similar to individual actions evaluation) and by considering social groups indicated by G-Act mAP2. For G-Act mAP2, we consider a true positive as a box for which the social group and the social activity labels are correctly predicted.   
\\
\noindent
{\bf Test-set Results.} In Tab.~\ref{tab:final}, we show that our suggested framework outperforms~\cite{ehsanpour2020joint}, on JRDB-Act test-set using the public detection provided in the JRDB benchmark~\cite{martin2019jrdb} obtained from Faster-RCNN~\cite{ren2016faster} in each task. It is worth noting that the performance of Faster-RCNN on JRDB-Act test-set is 52.2 mAP which shows the complexity of the dataset in the detection task. To study the effect of detection on the performance of each task, we utilized MMPAT~\cite{he2021know}, a better-performing detection on JRDB, with 68.1 mAP on test-set in Tab.~\ref{tab:final} and realized that more accurate detection boosts the grouping performance by a large margin. However, it performs almost on par with Faster-RCNN detection boxes on individual action and social activity detection tasks. This finding shows that understanding human actions in JRDB-Act is inherently complex due to the unique challenges in the data including robot motion and camera perspective. These challenges and results highlight the need of the existing research methods, including human activity detection frameworks, to support this new application in these type of environments which are underrepresented in existing datasets. In Tab.~\ref{tab:diff}, we further investigate the effect of the annotated difficulty tag for each provided label including social group, individual action and social activity labels in the evaluation of each task. As observed, using only easy labels indicated by E in evaluation, yields the best performance. Using easy and moderate tags~[E,M], also used in ablation studies, perform worse with a relatively small gap compared to~[E] and using all the labels with easy, moderate and difficult tags~[E,M,D] performs the worst with a large gap compared to~[E,M].\\
\noindent
{\bf Limitation and Discussion.} The model’s performance in the given tasks relies on the detector performance in predicting individuals’ boxes as well as the model’s performance in classification and clustering of the detected boxes. The current low action mAP in Tab.~\ref{Tab:A_ab} using groundtruth bounding boxes, evaluated on easy and moderate action labels as well as the negligible effect of utilising more accurate detected bounding boxes as validated in Tab.~\ref{tab:final}, show the inherent complexity and challenges of JRDB-Act in understanding human actions due to the motion of the robot, camera perspective, and highly unbalanced action distribution with different difficulty levels. Thus, this dataset may challenge the existing action localization frameworks, demanding further research in this direction to tackle the associated unique complexities. Moreover, JRDB-Act is a multi-modal dataset and provides annotation for 3D data which can potentially contribute to the overall performance of the tackled tasks. However, utilising 3D input can mainly contribute to the downstream tasks, \eg~detection, tracking and extracting more accurate geometrical features. A better detection in turn, results in higher social grouping performance as substantiated in Tab.~\ref{tab:final}. Exploring the 3D sensor modality data and investigating sensor data fusion strategies can be considered as potential future work.
\vspace{-0.7em}
\section{Conclusion}
\noindent 
Learning to recognise human actions and their social groups in an unconstrained environment including crowded scenarios, with potentially highly unbalanced human daily actions from a stream of sensory data captured from a mobile robot, remains a challenge, due to the lack of a reflective large-scale dataset. In this paper, we introduced JRDB-Act, a dataset captured from a moving social robot platform, including spatio-temporal individual action and social group annotations conducive to the task of simultaneously detection of social groups, individual actions and social activities. We also developed an end-to-end trainable pipeline to serve as a baseline to tackle this multi-task problem. We believe the dense annotations, and natural complexities of JRDB-Act pose new challenges for future research in the vision and robotics community.
{\small
\bibliographystyle{ieee_fullname}
\bibliography{egbib}
}
\clearpage
\section{Supplementary Material}
\begin{figure}[t]
  \centering
		\includegraphics[clip,width=1\linewidth]{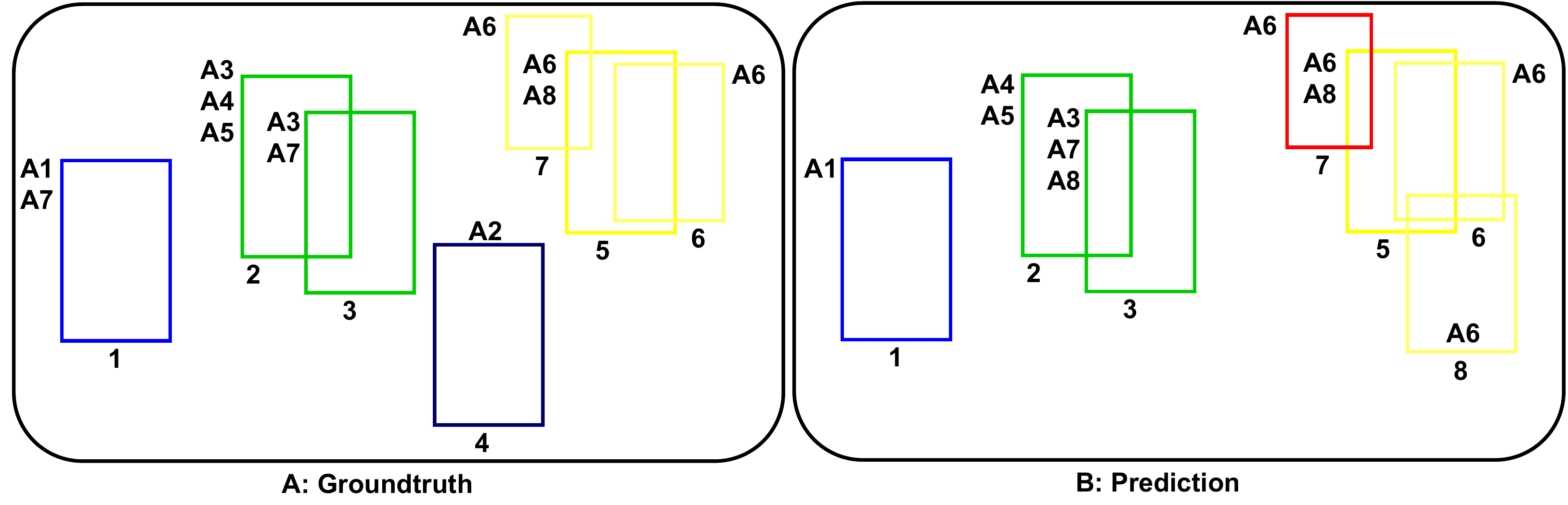}\vspace{-0.5em}
\caption{An example of the groundtruth and prediction scenarios for the three sub-tasks of individual action, social group and social activity detection. Matched groundtruth and detected bounding boxes are shown with similar number. The set of actions for each box is indicated by A$i$ next to it.}
\label{f_example}
\end{figure}

\noindent
{\bf 1.~Metric and Evaluation.}
We evaluate three sub-tasks on JRDB-Act namely, individual action, social group and social activity detection. We utilize the widely used Mean Average Precision~(mAP) on the key-frames of test-set. At inference, for each detected bounding box in the key-frame, our model predicts a set of individual action labels, a social group ID and we infer a set of social activity labels for that box by utilizing its social group ID and the predicted individual action labels of that group's members. The inferred social activity labels for all the bounding boxes of a social group, would be the common individual action labels of the members of that group,~\ie the actions performed by two or more people in the same group. Note that for singleton groups~(groups with one member), the social activity labels is identical to the person's individual actions. To explain the evaluation strategy of the three tasks, we show an example in Fig.~\ref{f_example}. We show bounding boxes by rectangles. Each bounding box has at least one individual action indicated by A$i$ and the color of bounding boxes indicate their social group ID. Consider A and B in Fig.~\ref{f_example} as the groundtruth and the prediction scenarios respectively.
\\
{\bf A.~Individual Action Evaluation. }Individuals' action detection evaluation is similar to the standard practice in~\cite{gu2018ava}. The true positive cases are (box1, A1), (box2, A4), (box2, A5), (box3, A3), (box3, A7), (box5, A6), (box5, A8), (box6, A6), (box7, A6) and the false negative cases are (box1, A7), (box2, A3), (box4, A2) and the false positive cases are (box3, A8), (box8, A6).
\\
{\bf B.~Social Grouping Evaluation. }For the social grouping evaluation, true positive cases are boxes 1,2,3,5,6, false negative case is box 4 and false positive cases are boxes 7,8. Social grouping performance is reported for social groups with 1, 2, 3, 4 and 5 or more members indicated by G1, G2, G3, G4 and G5$^+$ AP in Tab.~[2] and Tab.~[3] of the paper. Overall AP in these tables is the average of G1-G5$^+$.    
\\
{\bf C.~Social Activity Evaluation. }For the social activity evaluation, the inferred groundtruth social activity labels for the blue group is A1, A7, for the green group is A3, for the navy group is A2 and for the yellow group is A6.
Similarly, the predicted social activity labels for the blue group is A1, for the yellow group is A6 and for the red group is A6. For the green group we consider no predicted social activity label since none of the individual actions happening in that group is done by at least two people in the group. We assign the groundtruth and predicted per-group social activity labels to the members of that group. We evaluate social activity detection task in two ways. G-Act mAP1 evaluate the task by not considering the predicted social groups similar to the the individual action detection evaluation. In this scenario, true positive cases are (box1, A1), (box5, A6), (box6, A6) and {\bf (box7, A6)} although the social group of box 7 is wrongly predicted. False negative cases are (box1, A7),
(box2, A3), (box3, A3), (box4, A2) and the false positive cases are (box8, A6). On the other hand, G-Act mAP2 evaluate the task by considering the predicted social groups. In this scenario, {\bf (box7, A6)} is a false positive since its predicted group membership is not correct. Obviously, G-Act mAP2 is a stricter metric than G-Act mAP1 and clarifies the reason of lower performance in G-Act mAP2 compared to G-Act mAP1 in Tab.~[3] and Tab.~[4] of the paper.
\\
\noindent
{\bf 2.~Eigen-value based loss Proof.}
As stated in {\bf Learning Social Group Formation} of Sec.~[4] in the paper, the number of connected components~(social groups) in the groundtruth matrix $A$ is equal to the number of zero eigenvalues of its laplacian matrix ${L}$. Thus, we want the laplacian matrix of $A_{\theta}$ denoted by ${L}_{\theta}$ to have the same number of zero eigenvalues as in ${L}$.
If $e_{\theta}$ is an eigenvector of ${L}_{\theta}^{T}{L}_{\theta}$~(in order to ensure that the matrix is symmetric) with the eigenvalue $\lambda$, it satisfies ${L}_{\theta}^{T}{L}_{\theta}e_{\theta}=\lambda e_{\theta}$. Since $e^{T}_{\theta}e_{\theta}=1$~(eigenvectors have unit-norm), multiplying both sides of the equation by $e^{T}_{\theta}$ yields $e^{T}_{\theta}{L}_{\theta}^{T}{L}_{\theta}e=\lambda$ and we want to consider eigenvalues of zero~($\lambda=0$). Given the groundtruth eigenvector $e$ corresponding to the zero eigenvalue, we define the loss as \vspace{-.5em}
\begin{equation}
L_{eig}(\theta)=e^{T}{L}_{\theta}^{T}{L}_{\theta}e; \quad e^{T}{L}_{\theta}^{T}{L}_{\theta}e \geq 0 \vspace{-.5em}
\end{equation}
To avoid the trivial solution ${L}_{\theta}=0$, a second term is added to maximize the projection of data along the directions orthogonal to $e$.\vspace{-.5em}
\begin{equation}
L_{eig}(\theta)=e^{T}{L}_{\theta}^{T}{L}_{\theta}e - \alpha tr(\bar{{L}}_{\theta}^{T}\bar{{L}}_{\theta}) \vspace{-.5em}
\end{equation}
and finally for numerical stability the second term is bounded in the range $[0,1]$ as \vspace{-.5em}
\begin{equation}
L_{eig}(\theta)=e^{T}{L}_{\theta}^{T}{L}_{\theta}e+\alpha\exp(-\beta tr(\bar{{L}}_{\theta}^{T}\bar{{L}}_{\theta}))\vspace{-.5em} 
\end{equation}
This fully differentiable, eigendecomposition-free loss allows us to avoid performing eigendecomposition which suffers from the numerical instabilities of analytical differentiation.
\\
\noindent
{\bf 3.~JRDB-Act Action partitions. }As stated in {\bf Learning Actions} of Sec.~[4] in the paper, to improve the performance of individuals action detection at the presence of highly unbalanced action label distribution, we propsoe to utilise partitioning and balancing action loss  functions based on the occurring frequency of action classes in the dataset.
We utilise 3 cross entropy losses for 3 pose-based partitions:
{\bf [walking, standing, sitting], [cycling, going upstairs, bending], [going downstairs, skating, scootering, running]} and one binary cross entropy loss to learn whether there exists any interaction-based action and 3 more binary cross entropy losses for interaction-based partitions: {\bf [holding sth, listening to someone, talking to someone], [looking at robot, looking into sth, looking at sth, typing, interaction with door, eating sth], [talking on the phone, reading, pointing at sth, pushing, greeting gestures]}. Action labels are divided into disjoint partitions such that the occurring frequency of the most frequent action class is no more than 10 compared to the least frequent class in that partition.
\\
\noindent
{\bf 4.~Implementation Details.}
Our model's backbone~$f_{\theta}(x)$ is obtained from~\cite{ehsanpour2020joint} and here we elaborate its imp.~detail. We uses an I3D feature extractor which is initialized with Kintetics-400~\cite{kay2017kinetics} pre-trained model. We utilize ROI-Align with crop size of $5 \times 5$ on extracted feature-map from I3D. We perform self-attention on each individuals' feature map with query, key and value being different linear projections of individuals' feature map with output sizes being $1/8, 1/8, 1$ of the input size.
Individuals' feature maps obtained from the self attention module are fed into a single-layer, multi-head graph attention module with 8 heads with input, hidden and output dimention size of 1024 and droupout probability of 0.5 and $\alpha=0.2$ \cite{velikovi2017graph}. The rest of the imp.~detail is included in the paper.

\end{document}